\documentclass[journal]{IEEEtai}

\usepackage[colorlinks,urlcolor=blue,linkcolor=blue,citecolor=blue]{hyperref}
\usepackage{color,array}
\usepackage{graphicx}

\usepackage{cite}
\usepackage{amsmath,amssymb,amsfonts}
\usepackage{algorithmic}
\usepackage{graphicx}
\usepackage{textcomp}
\usepackage{xcolor}
\usepackage{soul}
\usepackage{url}
\usepackage{hyperref}
\usepackage{booktabs}

\begin{document}

\title{Variational Graph Convolutional Neural Networks}

\author{Illia Oleksiienko, Juho Kanniainen, and Alexandros Iosifidis, \IEEEmembership{Senior Member, IEEE}
\thanks{The research received funding from the Independent Research Fund Denmark project DeepFINA (grant ID 10.46540/3105-00031B).}
\thanks{Illia Oleksiienko is with the Department of Electrical and Computer Engineering, Aarhus University, Denmark (e-mail: io@ece.au.dk).}
\thanks{Juho Kanniainen is with the Faculty of Information Technology and Communication Sciences, Tampere University, Finland (e-mail: juho.kanniainen@tuni.fi).}
\thanks{Alexandros Iosifidis is with the Faculty of Information Technology and Communication Sciences, Tampere University, Finland (e-mail: alexandros.iosifidis@tuni.fi).}}

\markboth{Journal of IEEE Transactions on Artificial Intelligence, Vol. 00, No. 0, Month 2025}
{Illia Oleksiienko \MakeLowercase{\textit{et al.}}: Variational Graph Convolutional Neural Networks}


\maketitle

\begin{abstract}
Estimation of model uncertainty can help improve the explainability of Graph Convolutional Networks and the accuracy of the models at the same time.
Uncertainty can also be used in critical applications to verify the results of the model by an expert or additional models.
In this paper, we propose Variational Neural Network versions of spatial and spatio-temporal Graph Convolutional Networks.
We estimate uncertainty in both outputs and layer-wise attentions of the models, which has the potential for improving model explainability. We showcase the benefits of these models in the social trading analysis and the skeleton-based human action recognition tasks on the Finnish board membership, NTU-60, NTU-120 and Kinetics datasets, where we show improvement in model accuracy in addition to estimated model uncertainties.
\end{abstract}



\section{Introduction}
The ability of Graph Convolutional Networks (GCNs) \cite{kipf2017gcn, velichkovic2018gat} to capture local and global information, as well as the effective use of computational resources, make them a favorable option for graph-related problems, both spatial and spatio-temporal ones, such as Social Media analysis \cite{jin2022gcn_social, bian2020bigcn_social}, (social) trading behavior analysis \cite{baltakys2023ego, qian2024gnn_stock}, skeleton-based human action recognition \cite{yan2018stgcn, shi2019agcn, heidari2021tagcn} and chemical compound analysis \cite{zeng2024gnn_ddas, elbasani2022_gcrnn}. Spatial tasks require analysis of a single instance of graph data, while spatio-temporal tasks consider also the changes to the graph that occur in time.

Depending on the application of GCNs, actions taken as a result of the analysis by the model can be costly or even dangerous if the model is mistaken and there is no process in place to mitigate such mistakes.
For example, if an investor's trading activity is flagged as suspicious by a GCN, authorities may need to investigate suspicious actions, which, in the case of a mistake, will either cost financial resources or, in the worst case, could lead to penalty to an innocent person, if too much trust is placed on the model.
To address such problems, models that estimate the uncertainty in their predictions could be utilized.

Uncertainty estimation methods attempt to provide an uncertainty value, in addition to the output of the model, which represents epistemic, aleatoric, or a combined uncertainty in the generated output.
Uncertainty estimation in neural network predictions can better indicate if the model output should be used as is or if any additional actions should be performed.
For example, when human action recognition is used within a human-robot interaction functionality, uncertainty estimation can indicate when the model is not able to classify the action performed by the human with good certainty and additional confirmation or stopping the process is required to avoid dangerous situations.
In social trading analysis, the model uncertainty can indicate which trades should be investigated further.
In addition to this, uncertainty estimation in the model attentions can be used to improve model explainability and guide the training and design process for each particular application.

In this paper, we propose a Variational Neural Network (VNN) \cite{oleksiienko2023vnn} version of Graph Convolutional Networks with different architectures, including GCN \cite{kipf2017gcn} and GAT \cite{velichkovic2018gat} for spatial models, and ST-GCN \cite{yan2018stgcn} and AGCN \cite{shi2019agcn} for spatio-temporal models.
We showcase the benefits of the proposed spatial models in the social trading analysis task and those of the proposed spatio-temporal models in skeleton-based human action recognition tasks.
In experiments, we show that the variational versions of the models provide an improvement in the classification performance while also providing the uncertainty in both outputs and attentions of the models.

\section{Related Work}
Uncertainty estimation in neural networks can be done in four main ways \cite{gawlikowski2021uncertaintyindl}.
Deterministic methods \cite{2018evidentialdl, zhong2020uavoxel} are based on the use of a single model that either regresses the uncertainty in a separate branch, or computes some properties of the output.
Bayesian Neural Networks (BNNs) \cite{blundell2015weight, magris2023BNNsurvey} utilize a distribution over weights to create a stochastic model that, by considering multiple weight samples, can estimate model uncertainty via Monte Carlo integration of the outputs from different sampled models.
Ensemble Methods \cite{osband2018randomized, valdenegro2019subens, oleksiienko2023lens} are a specific case of BNNs where the distribution is categorical, resulting in a set of models that are trained in parallel and used together to compute the total output and uncertainty.
Test-Time Data Augmentation methods \cite{wang2018barintumortesttimeaug, wang2019aleamedical, kandel2021testtime} change the input to the static model by applying different augmentation and analyze the difference in outputs to estimate model uncertainty.
In contrast to BNNs and Ensemble Methods, Variational Neural Networks (VNNs) \cite{oleksiienko2023vnn, oleksiienko2022vnntorchjax} use only one set of weights, but the inputs are processed to parametrize a Gaussian distribution for each layer.
VNNs sample the output of the layer from the generated distribution, introducing stochasticity to the model.

Uncertainty estimation in GNNs is surveyed in \cite{wang2024uncertaintygnnsurvey}.
The authors classify the methods in the first three aforementioned categories, excluding the Test-Time Data Augmentation methods which are also present for the GNNs, including GraphPatcher \cite{ju2023graphpatcher} and a method based on test-time augmentation \cite{bo2021testtimesocial}.

Bayesian approaches, in general, provide better statistical models than non-Bayesian methods \cite{osband2021epistemic}, meaning that Deterministic and Test-time augmentation methods can be statistically improved by considering their Bayesian counterparts.
However, Bayesian methods are usually harder to implement, and they require more resources than deterministic methods to estimate uncertainty.
Among BNNs, Monte Carlo Dropout (MCD) \cite{2016dropout} provides the worst quality of uncertainty \cite{osband2021epistemic, oleksiienko2023vnn}, but it is widely used due to the ease of application to an existing model.
The popularity of MCD is also confirmed by the aforementioned survey, as they are widely used for GNN-based uncertainty estimation \cite{zhang2019mcgcn, hasanzadeh2020mcgcn, rong2020dropedge}.
The proposed framework for Variational GCNs allows using a better quality of uncertainty \cite{oleksiienko2023vnn} than the popular MCD without the sacrifices in the ease of use \cite{oleksiienko2022vnntorchjax}.

\section{Background}
The input to a GNN is usually represented as a feature matrix $S \in \mathbb{R}^{C \times N}$ together with an Adjacency matrix $A \in \mathbb{R}^{N \times N}$, where $C$ is the number of features (or channels) and $N$ is the number of nodes.
Graph Convolutional Networks (GCNs) \cite{kipf2017gcn} are a type of GNNs that consist of multiple Graph Convolutional Layers $\Pi(\cdot)$. The input features to the $i$-th Graph Convolutional Layer $S^i$ are transformed through multiplication with a weight matrix $W$ and a normalized Adjacency matrix $\hat{A}$ as follows: 
\begin{align}
\begin{split}
    &\Pi(S^i) = \rho(\hat{A} S^i W),\\
    &\hat{A} = D^{-0.5} (A + I) D^{-0.5},
\end{split}
\end{align}
where $D$ is the graph Degree matrix and $\rho(\cdot)$ is an activation function applied in an element-wise manner to its input.

Graph Attention Networks (GAT) \cite{velichkovic2018gat} consider the initial Adjacency matrix not as a strict limitation of information propagation, but combine it with a learned and data-dependent attention matrix for allowing problem-specific graph node connections to be learned, targeting improving the model generalization ability.
Each Graph Attention Layer $\Theta(\cdot)$ of a GAT network creates an attention matrix $\Lambda$ and combines it with the Adjacency matrix $A$ and the input matrix $S^i$ as follows:
\begin{align}
\begin{split}
    &\Theta(S^i) = \rho(\Lambda) \hat{S^i},\\
    &\hat{S^i} = S^i W, \\
    &\Lambda = \rho_\Lambda(\lambda_s \lambda_d^T) \odot (1 - A),\\
    &\lambda_s = \text{tanh}(\hat{S^i}) w_s,\\
    &\lambda_d = \text{tanh}(\hat{S^i}) w_d,\\
\end{split}
\end{align}
where $\rho(\cdot)$ is the activation function for the attention matrix after fusion with the Adjacency matrix $A$, $\hat{S^i}$ are the transformed input features, $\rho_\Lambda(\cdot)$ is an activation function for the attention matrix before fusion with the Adjacency matrix A, $\odot$ is the Hadamard product, $\text{tanh}(\cdot)$ is the hyperbolic tangent function, and $w_s$ and $w_d$ are the so-called source and destination attention weights.

\begin{figure}[!b]
\centering
    \includegraphics[width=0.9\linewidth]{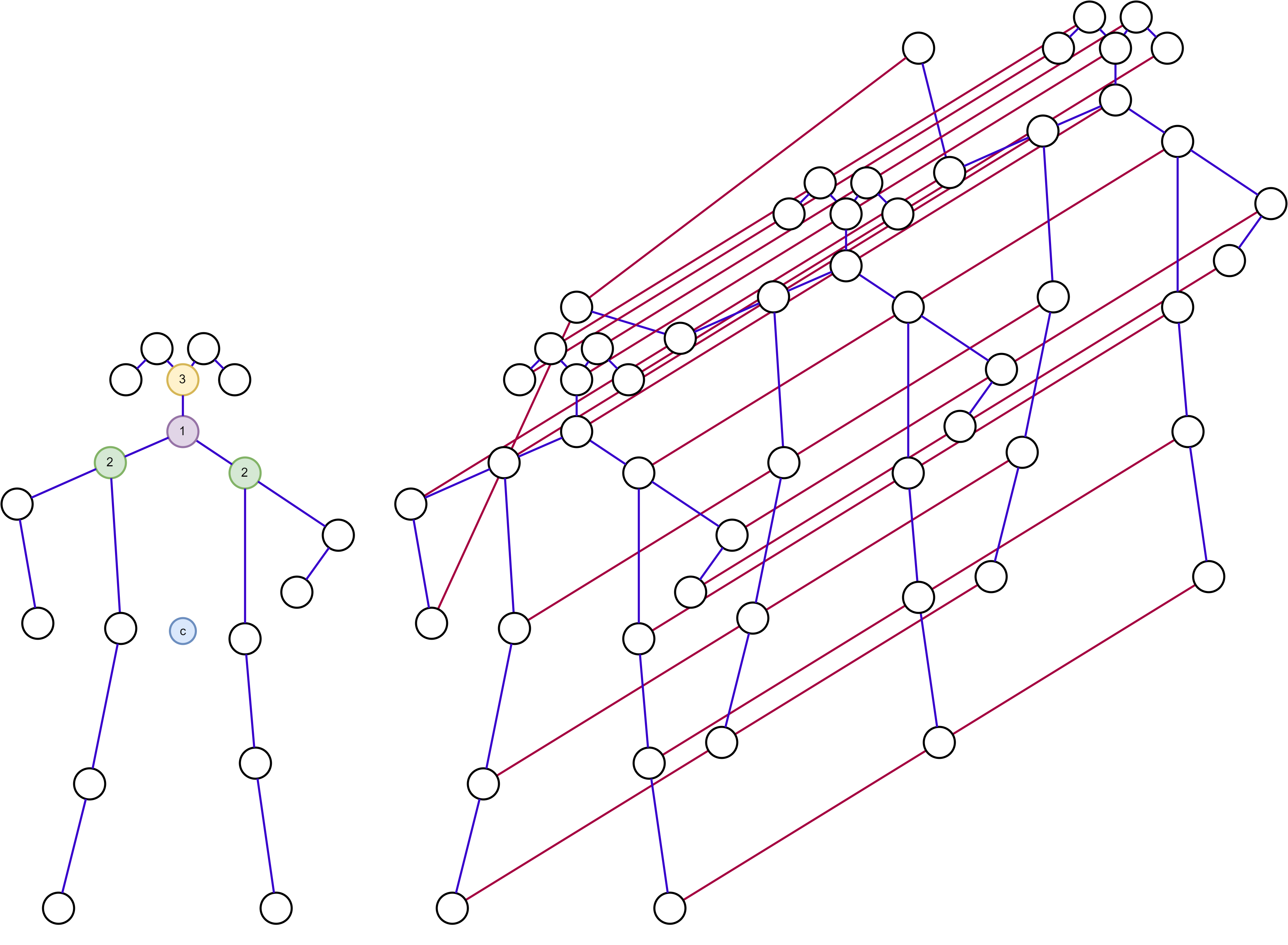}
    \caption{An example of spatio-temporal human body poses based on human body skeletons coming from the Kinetics \cite{will2017kinetics400} dataset. A single skeleton (left) and a temporal set of skeletons (right) are present in the figure. The purple connections represent spatial edges between nodes, and the red connections represent temporal edges. For each skeleton node, its neighbors can be classified into three groups: (1) ego joint, (2) joints that are closer to the center joint (c), and (3) joints that are farther from the center joint.}
    \label{fig:skeleton-example}
\end{figure}

Spatio-Temporal Graph Convolutional Networks \cite{yan2018stgcn, shi2019agcn} process a sequence of graphs in a (2+1)-D manner \cite{tran20182p1dconv}, where each of the spatial 2D inputs is processed independently, and then features collected for all time instances are aggregated over the temporal dimension.
The spatio-temporal data processing in ST-GCN is done over multiple layers, each of which employs a GCN layer processing each spatial graph followed by the temporal processing done by an 1D convolution.

The input sequence is represented as a 3D tensor $S \in \mathbb{R}^{C \times T \times N}$, where $C$ is the number of features (channels) for each node, $T$ is the number of frames and $N$ represents the number of nodes in a graph. 
The graph connections are stored in a binary Adjacency matrix $A \in \mathbb{R}^{N \times N}$. Depending on the task to be solved, multiple task-specific Adjacency matrices can be employed. Both ST-GCN \cite{yan2018stgcn} and AGCN \cite{shi2019agcn}, originally proposed for the task of skeleton-based human action recognition, employ three Adjacency matrices $A_p, \:p \in \{1,2,3\}$, each of which encodes $p$-order node connections. As shown in Figure \ref{fig:skeleton-example}, the first Adjacency matrix encodes self-connections, the second one encodes connections to nodes closer to the geometrical graph center, and the third one encodes connections to nodes farther from the center.
Each Adjacency matrix is then normalized as follows: 
\begin{equation}
    \hat{A}_p = D_p^{-0.5} (A_p + I) D_p^{-0.5},
\end{equation}
where $D_p$ represents the corresponding graph Degree matrix.

Spatio-temporal Graph Convolutional Networks consist of multiple blocks $\Gamma(\cdot)$. These blocks transform the input features $S^i$ into output features $S^{i+1}$ as follows:
\begin{align}
\begin{split}
    &\Gamma(S^{i}) = \rho\left(\Xi(S^{i}) + \text{BN}(\text{TC}(G(S^i))\right),\\
    &G(S^i) = \sum_p(\hat{A}_p \circ M_p) S^{i} W_p,\\
    \label{eq:general-stgcn}
\end{split}
\end{align}
where $\rho(\cdot)$ is the ReLU activation function, $\text{BN}(\cdot)$ is a batch normalization function, $\text{TC}(\cdot)$ is a temporal convolution function, $\circ$ is an attention fusion function which can be either an element-wise multiplication or a matrix addition, $W_p$ is a learnable weight matrix corresponding to the $p$-th order connections, and $M_p$ is the corresponding attention matrix which can be either learnable or a computed one. The $\Xi(\cdot)$ function implements skip-connections by ensuring that the number of channels of the input and the output are compatible as follows:
\begin{equation}
    \Xi(S^{i}) = 
    \begin{cases}
        S^{i}, & \text{if } C^{i} = C^{i+1}, \\
        S^{i} W_\xi, & \text{otherwise},
    \end{cases}\\
\end{equation}
where $W_\xi$ is a learnable matrix that transforms input features to have $C^{i+1}$ channels, which is the number of channels in the output value $S^{i+1}$.

ST-GCN \cite{yan2018stgcn} and AGCN \cite{shi2019agcn} are specific implementations of the above-described network structure having different attention mechanisms based on the attention fusion operator $\circ$ in Equation (\ref{eq:general-stgcn}).
ST-GCN uses the Hadamard product to combine the Adjacency matrices as follows:
\begin{align}
\begin{split}
    &\Gamma_{\text{stgcn}}(S^{i}) = \rho(\Xi(S^{i}) + \text{BN}(\text{TC}(G_{\text{stgcn}}(S^i)))),\\
    &G_{\text{stgcn}}(S^i) = \sum_p(\hat{A}_p \odot M_p) S^{i} W_p.
\end{split}
\end{align}
AGCN splits the computed attention matrix $M_p$ for each of the partitions into the learned attention matrix $W^M_p$ and the computed attention matrix $B_p$ as follows:
\begin{align}
\begin{split}
    &\Gamma_{\text{agcn}}(S^{i}) = \rho(\Xi(S^{i}) + \text{BN}(\text{TC}(G_{\text{agcn}}(S^i)))),\\
    &G_{\text{agcn}}(S^i) = S^i + \text{BN}(\sum_p Z_p),\\
    &Z_p = \text{conv}(S^{i} (\hat{A}_p + M_p), W_{z,p}),\\
    &M_p = W^M_p + B_p,\\ 
    &B_p = \frac{\text{softmax}(B_{1,p} B_{2,p})}{N},\\
    &B_{q,p} = \text{conv}(S^i, W_{q,p}) \:\: \forall q \in \{1,2\},\\
\end{split}
\end{align}
where $\text{conv}(\cdot, W)$ is a 2D convolution function parametrized by weights $W$. $W_{z,p}$, $W_{1,p}$ and $W_{2, p}$ are the convolution parameters for the feature combination part $Z_p$ and attention matrix generation parts $B_{1,p}$ and $B_{2,p}$, respectively. $\text{softmax}(\cdot)$ is the softmax function, and $N$ is the number of nodes in the graph.

\section{Variational Graph Convolutional Networks}
We propose the Variational Graph Convolutional Networks, which include the Variational GCN (VGCN) and Variational GAT (VGAT) networks for spatial tasks, and the Variational Spatio-temporal Graph Convolutional Networks (VST-GCNs) for spatio-temporal tasks. This is done by implementing Variational Neural Network versions of the aforementioned networks.

The variational version of GCN consists of multiple Variational GCN layers, each of which consists of two graph convolutional sub-layers that compute parameters for a Gaussian distribution and the (possibly activated) sampled values from this distribution are used as the output of the layer:
\begin{align}
\begin{split}
    &\Pi_{\text{vgcn}}(S^i) = \rho^{\mathcal{N}}(\tilde{\Pi}(S^i)),\\
    &\dot{\Pi}^{\nu}(S^i) = \hat{A} S^i W^{\nu},\\
    &\tilde{\Pi}(S^i) \sim \mathcal{N}\Big(\rho^{\mu}(\dot{\Pi}^{\mu}(S^i)), \rho^{\sigma}(\dot{\Pi}^{\sigma}(S^i))\Big),\\
\end{split}
\end{align}
where $\nu \in \{\mu, \sigma\}$ and $\sim$ is the sampling operator, $\mathcal{N}(\cdot, \cdot)$ is a Gaussian distribution function. $\rho^{\mathcal{N}}(\cdot)$, $\rho^{\mu}(\cdot)$ and $\rho^{\sigma}(\cdot)$ are the activation functions for the outputs, means and variances, respectively.
The activation functions can be either the identity function or a nonlinear activation function such as ReLU.

\begin{figure}[!ht]
\centering
    \includegraphics[width=\linewidth]{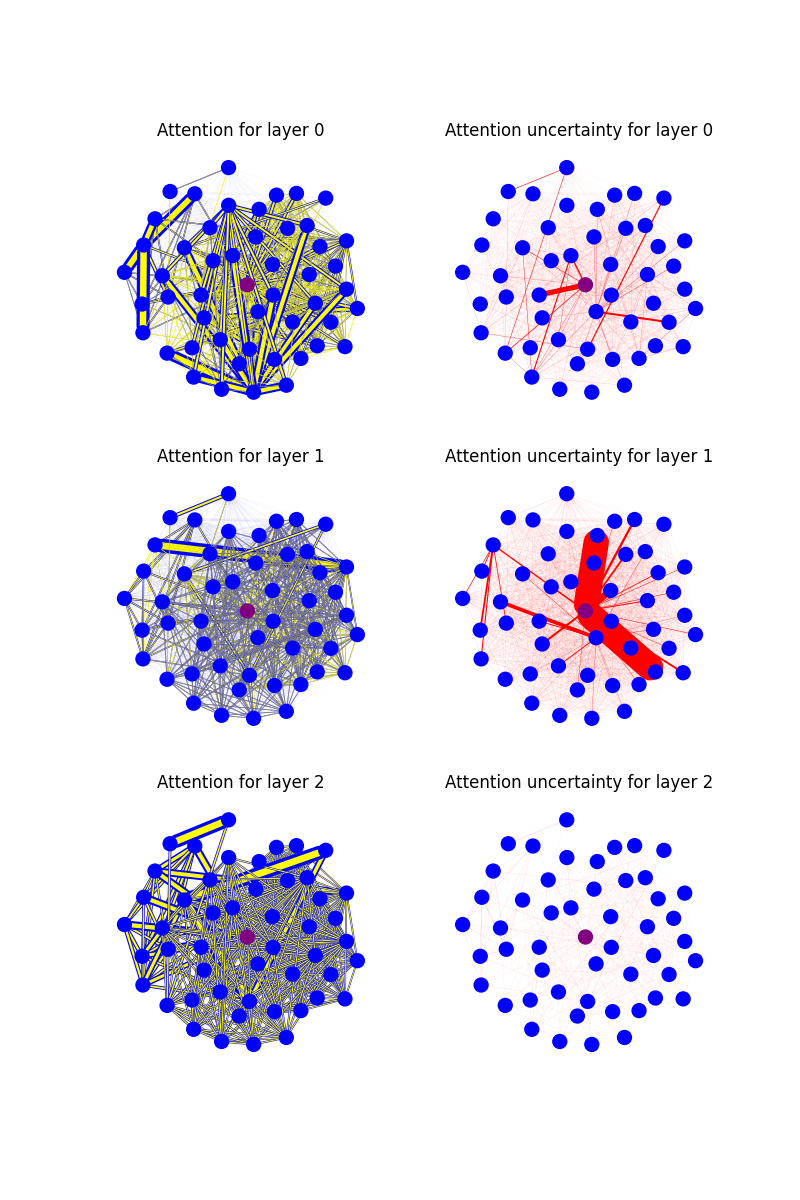}
    \caption{An example of layer-wise attention graphs in a single investor ego graph with corresponding uncertainties. The width of the blue lines represents the value of attention between the pair of investors, while the yellow lines represent the original binary adjacency matrix. The width of the yellow lines is adjusted to the corresponding attention value for visual purposes and does not define the weight of the connection in the adjacency matrix. The width of the red lines represents the uncertainty in the attention value, relative to the attention value itself.}
    \label{fig:attention-graph-insider}
\end{figure}

The VGAT network follows the same principle as VGCN. The computations of the attention matrix $\Lambda$ are done for both sub-layers as follows:
\begin{align}
\begin{split}
    &\Theta_{\text{vgat}}(S^i) = \rho^{\mathcal{N}}(\tilde{\Theta}(S^i)),\\
    &\dot{\Theta}^{\nu}(S^i) = \rho(\Lambda^{\nu}) \hat{S_{\nu}^i},\\
    &\hat{S_{\nu}^i} = S^i W^{\nu}, \\
    &\Lambda^{\nu} = \rho_\Lambda(\lambda_s^{\nu} {\lambda_d^{\nu}}^T) \odot (1 - A),\\
    &\lambda_s = \text{tanh}(\hat{S^i_{\nu}}) w_s^{\nu},\\
    &\lambda_d = \text{tanh}(\hat{S^i_{\nu}}) w_d^{\nu},\\
    &\tilde{\Theta}(S^i) \sim \mathcal{N}\Big(\rho^{\mu}(\dot{\Theta}^{\mu}(S^i)), \rho^{\sigma}(\dot{\Theta}^{\sigma}(S^i))\Big),\\
\end{split}
\end{align}
where $\nu \in \{\mu, \sigma\}$.
This allows us to not only analyze the outputs of the model, but also to define the so-called \textit{uncertain attentions} in the same way as uncertain outputs:
\begin{align}
\begin{split}
    &\tilde{\Lambda} \sim \mathcal{N}\Big(\Lambda^{\mu}, \Lambda^{\sigma}\Big).\label{eq:uncertain_attentions}
\end{split}
\end{align}
The mean and variance of uncertain attentions can be obtained by applying Monte Carlo integration over multiple uses of the model, resulting in $\tilde{\Lambda}^{\mu}$ and $\tilde{\Lambda}^{\sigma}$ as the overall expectation and uncertainty of the attention at each layer of the model.
However, we can avoid this step if, instead of sampling the uncertain attentions from the Gaussian distribution, we decouple it back into mean and variance of the attention matrix, but this process is only possible to do when attention does not depend on inputs. The mean and variance of the uncertain attentions of VGAT cannot be computed without the Monte Carlo process since both $\Lambda^{\mu}$ and $\Lambda^{\sigma}$ are computed based on $S^i$, which is in most cases sampled from a Gaussian distribution of the previous layer, resulting in different values of $\Lambda^{\mu}$ and $\Lambda^{\sigma}$ through different iterations of applying the model to the same input.
Examples of attention means and uncertainties obtained for the social trading analysis task where a graph of investors in the neighborhood of a target investor is analyzed by a 3-layer VGAT model can be seen in Figure \ref{fig:attention-graph-insider}.
Attentions for each of the layers are shown as expectation and uncertainty graphs.
\begin{figure}[!ht]
\centering
    \includegraphics[width=0.8\linewidth]{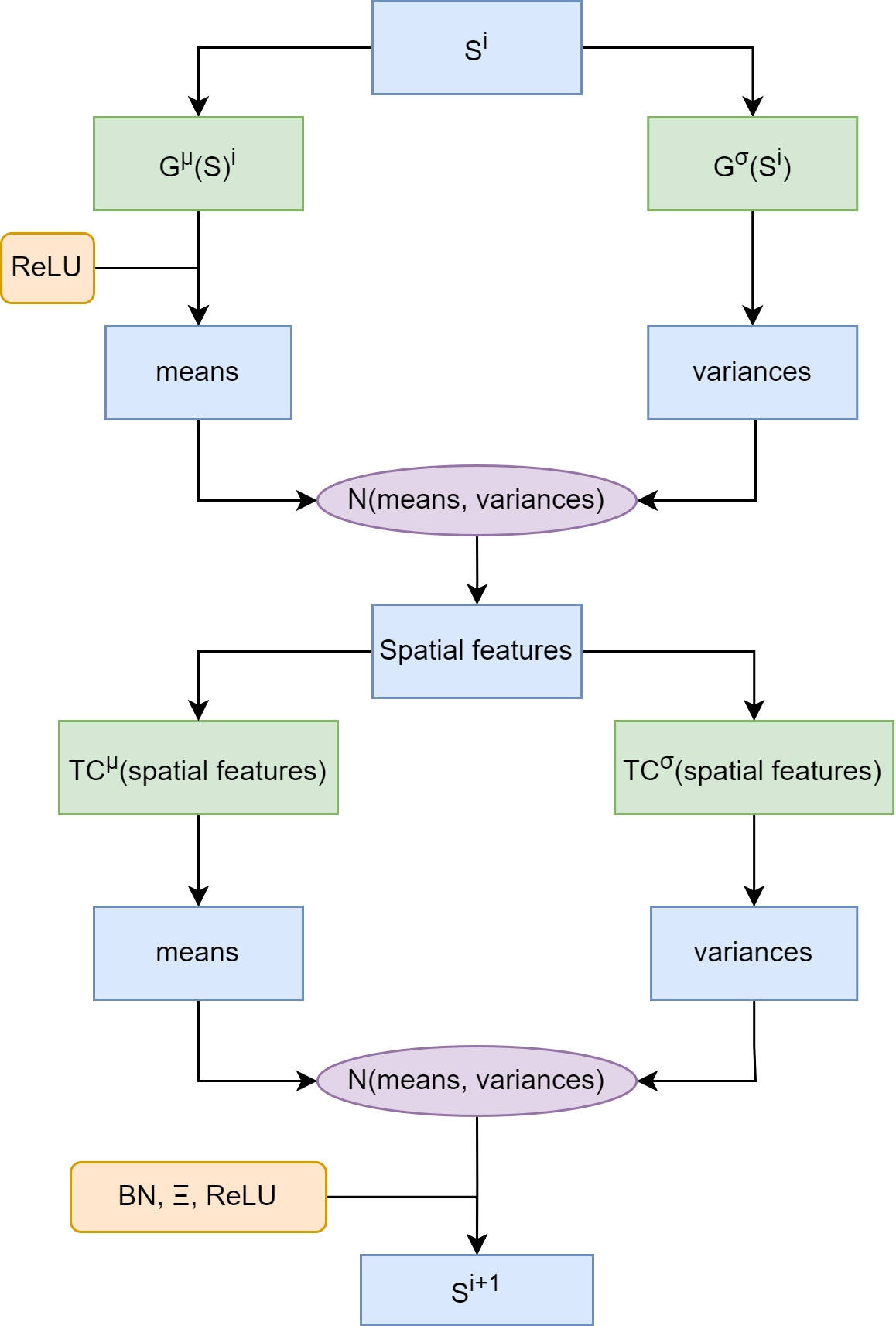}
    \caption{Structure of a Variational ST-GCN block. The input features $S^i$ at layer $i$ are processed by two spatial GCN sub-layers $\Gamma_\mu(\cdot)$ and $\Gamma_\sigma(\cdot)$ to create mean and variance parameters of the corresponding Gaussian distribution. The sampled outputs represent the spatial features, which are then processed by two temporal convolution sub-layers $\text{TC}_\mu(\cdot)$ and $\text{TC}_\sigma(\cdot)$, outputs of which are used to create another Gaussian distribution. Batch normalization, activation and residual transformation function are applied to the sampled values to create the final block outputs.}
    \label{fig:vstgcn-block}
\end{figure}

Both Variational ST-GCN and Variational AGCN are defined through a Variational ST-GCN block, which is shown in Figure \ref{fig:vstgcn-block}. The Variational ST-GCN block implements a Variational Layer on both spatial and temporal parts of the block.
This is done by first processing the input $S^i$ with two spatial sublayers $G^{\mu}(\cdot)$ and $G^{\sigma}(\cdot)$, which have the same structure but are parametrized with different weights. The output of the $G^{\mu}(S^i)$ part is activated by a ReLU activation function, and both sub-layer outputs are used to parametrize a Gaussian distribution. The spatial outputs are sampled from the generated Gaussian distribution and are then used as inputs to the temporal variational part. This part applies two temporal convolutions $\text{TC}^{\mu}(\cdot)$ and $\text{TC}^{\mu}(\cdot)$ to parametrize the final Gaussian distribution, outputs of which are processed by batch normalization, $\Xi(\cdot)$, and activation function, following Equation (\ref{eq:general-stgcn}).
The mathematical definition of the VSTGCN block $\Gamma_{\text{vstgcn}}(\cdot)$ is:
\begin{align}
\begin{split}
    &\Gamma_{\text{vstgcn}}(S^{i}) = \rho(\Xi(S^{i}) + \text{BN}(\tilde{\text{TC}}(\tilde{G}_{\text{stgcn}}(S^i))),\\
    &G_{\text{stgcn}}^{\nu}(S^i) = \sum_p(\hat{A}_p \odot M^{\nu}_p) S^{i} W^{\nu}_p,\\
    &\tilde{G}_{\text{stgcn}}(S^i) \sim \mathcal{N}\Big(\rho^{\mu}(G^{\mu}_{\text{stgcn}}(S^i)), G^{\sigma}_{\text{stgcn}}(S^i)\Big),\\
    &\tilde{\text{TC}}(X) \sim \mathcal{N}\Big(\text{TC}^{\mu}(X), \text{TC}^{\sigma}(X)\Big),
\end{split}
\end{align}
where $\nu \in \{\mu, \sigma\}$, $\text{TC}^{\mu}(\cdot)$ and $\text{TC}^{\sigma}(\cdot)$ are the temporal convolution functions parametrized by different weights, and $\rho^\mu(\cdot)$ is the activation function for mean values of the spatial graph convolution.
The VAGCN block $\Gamma_{\text{vagcn}}(\cdot)$ can be defined in a similar manner:
\begin{align}
\begin{split}
    &\Gamma_{\text{vagcn}}(S^{i}) = \rho(\Xi(S^{i}) + \text{BN}(\tilde{\text{TC}}(\tilde{G}_{\text{agcn}}(S^i))),\\
    &\forall \nu \in \{\mu, \sigma\},\\
    &G^{\nu}_{\text{agcn}}(S^{i}) = S^i + \text{BN}(\sum_p Z^{\nu}_p),\\
    &Z^{\nu}_p = \text{conv}(S^{i} (\hat{A}^{\nu}_p + M^{\nu}_p), W^{\nu}_{z,p}),\\
    &M^{\nu}_p = W^{M, \nu}_p + B^{\nu}_p,\\ 
    &B^{\nu}_p = \frac{\text{softmax}(B^{\nu}_{1,p} B^{\nu}_{2,p})}{N},\\
    &B^{\nu}_{i,p} = \text{conv}(S^i, W^{\nu}_{i,p}) \:\: \forall i \in \{1,2\}, \\
    &\tilde{G}_{\text{agcn}}(S^i) \sim \mathcal{N}\Big(\rho^{\mu}(G^{\mu}_{\text{agcn}}(S^i)), G^{\sigma}_{\text{agcn}}(S^i)\Big),\\
    &\tilde{\text{TC}}(X) \sim \mathcal{N}\Big(\text{TC}^{\mu}(X), \text{TC}^{\sigma}(X)\Big).
\end{split}
\end{align}

Similarly to VGAT, we can compute attention uncertainty for both VSTGCN and VAGCN models.
VSTGCN model attentions do not depend on the inputs and are trained, so we can directly get the mean and the variance of the attentions from the model weights, while the VAGCN attentions depend on the input and thus should be computed identically to VGAT.
Additionally, for both VGAT and VAGCN we can compare the \textit{raw} attentions which are stored in the weights and the \textit{final} attention values which are computed as the combination of the \textit{raw} attention and inputs.

\subsection{Uncertainty-Aware models}
The attention uncertainty estimation for Variational GCN models can be further utilized to create Uncertainty-Aware Variational GCN models.
We implement two methods that utilize the uncertainty in attentions of the Variational GCNs.
Both methods are based on the uncertain attentions obtained using Equation (\ref{eq:uncertain_attentions}), and the Monte Carlo integration process that combines uncertain attentions from multiple samples of the network, i.e., attention mean $\tilde{\Lambda}^\mu$ and attention variance $\tilde{\Lambda}^\sigma$.
Then, the attention mean and variance are filtered based on the attention variance values:
\begin{align}
\begin{split}
    \forall k,q, \:\:\:\: \tilde{\Lambda}^\mu_{\text{filtered}}[k,q] =
    \begin{cases}
        \tilde{\Lambda}^\mu[k,q], \:\: \text{if} \:\:\tilde{\Lambda}^\sigma[k,q] \leq l \tilde{\Lambda}^\mu[k,q],\\
        p, \:\: \text{otherwise},
    \end{cases}\\
    \forall k,q, \:\:\:\: \tilde{\Lambda}^\sigma_{\text{filtered}}[k,q] =
    \begin{cases}
        \tilde{\Lambda}^\sigma[k,q], \:\: \text{if} \:\:\tilde{\Lambda}^\mu[k,q] \leq l \tilde{\Lambda}^\sigma[k,q],\\
        0, \:\: \text{otherwise},
    \end{cases}
    \label{eq:filtered_attention}
\end{split}
\end{align}
where $X[k,q]$ is the value of the matrix $X$ at position $(k,q)$, $l$ is the attention filter limit, which defines how high the uncertainty in the specific attention value should be to filter it out, and $p$ is the replacement value set to a low number such as $0$ or $0.01$.

The filtered matrix can be utilized in two ways. The \textit{Early Attention} approach changes the formulation of a Variational GCN model to combine features and attentions from mean and variance sub-layers early, and then process the combined result. The Uncertainty-Aware Early Attention VGAT (UA-EA-VGAT) model is formulated as follows:
\begin{align}
\begin{split}
    &\Theta_{\text{uaeavgat}}(S^i) = \Omega(\hat{S^i}, \tilde{\Lambda}^\mu_{\text{filtered}}),\\
    & \Omega(\hat{S^i}, \tilde{\Lambda}^\mu_{\text{filtered}}) = \rho(\tilde{\Lambda}^\mu_{\text{filtered}}) \rho^{\mathcal{N}}(\hat{S^i}),\\
    &\forall \nu \in \{\mu, \sigma\},\\
    &\hat{S_{\nu}^i} = S^i W^{\nu}, \\
    &\Lambda^{\nu} = \rho_\Lambda(\lambda_s^{\nu} {\lambda_d^{\nu}}^T) \odot (1 - A),\\
    &\lambda_s = \text{tanh}(\hat{S^i_{\nu}}) w_s^{\nu},\\
    &\lambda_d = \text{tanh}(\hat{S^i_{\nu}}) w_d^{\nu},\\
    &\hat{S^i} \sim \mathcal{N}\Big(\rho^{\mu}(\hat{S^i_{\mu}}), \rho^{\sigma}(\hat{S^i_{\sigma}})\Big),\\
\end{split}
\end{align}
where the filtered attention matrix $\tilde{\Lambda}^\mu_{\text{filtered}}$ is computed following Equations (\ref{eq:uncertain_attentions}) and (\ref{eq:filtered_attention}). $\Omega(\cdot, \cdot)$ is the output step function that combines the sampled features $\hat{S^i}$ and the filtered attention matrix $\tilde{\Lambda}^\mu_{\text{filtered}}$.

This formulation combines the mean and variance branches of a VGAT layer early, directly producing two uncertain outputs, i.e., features $\hat{S^i}$ and attentions $\tilde{\Lambda}^\mu$, which are then combined in a classical manner through the output step function.
Such change forces the distribution of the output to be different from the VGAT, even if we omit the filtering process.
Considering the reparametrization trick \cite{kingma2014autoencoding}, we can express the output of a VGAT layer as
\begin{align}
\begin{split}
&\Theta_{\text{vgat}}(S^i) = \rho^{\mathcal{N}}(\rho^{\mu}(\rho(\Lambda^{\mu}) \hat{S_{\mu}^i}) + \rho^{\sigma}(\rho(\Lambda^{\sigma}) \hat{S_{\sigma}^i})^{\frac{1}{2}} \epsilon,\\
& \epsilon \sim \mathcal{N}(0, I),
\end{split}
\end{align}
and if we omit the filtering process and propagate attention matrix directly, the output of the Early Attention VGAT layer is parametrized by two distributions:
\begin{align}
\begin{split}
&\Theta_{\text{uaeavgat}}(S^i) = \\
& =\rho(\Lambda^\mu + (\Lambda^\sigma)^\frac{1}{2} \epsilon_\Lambda)\rho_{\mathcal{N}}(\rho^\mu(\hat{S^i_\mu}) + \rho^\sigma(\hat{S^i_\sigma})^\frac{1}{2} \epsilon_S), \\
& \epsilon_\Lambda \sim \mathcal{N}(0, I), \\
& \epsilon_S \sim \mathcal{N}(0, I),
\end{split}
\end{align}
which means that a pretrained VGAT model cannot be converted directly to the UA-EA-VGAT model and such models should be trained from scratch.

The second approach, namely the \textit{Fully Monte Carlo Integrated} (FMCI) method, keeps the output distribution of the original model the same, and therefore can be used without retraining the model.
To create the Uncertainty-Aware FMCI VGAT model, we want to combine attentions from different samples in such a way that, when the filtered attention replaces the original attention, the model can proceed in the same manner as if the attentions are unchanged:
\begin{align}
\begin{split}
    &\Theta_{\text{uafmcivgat}}(S^i) = \rho^{\mathcal{N}}(\tilde{\Theta}(S^i)),\\
    &\forall \nu \in \{\mu, \sigma\},\\
    &\dot{\Theta}^{\nu}(S^i) = \rho(\tilde\Lambda^{\nu}_\text{filtered}) \hat{S_{\nu}^i},\\
    &\hat{S_{\nu}^i} = S^i W^{\nu}, \\
    &\Lambda^{\nu} = \rho_\Lambda(\lambda_s^{\nu} {\lambda_d^{\nu}}^T) \odot (1 - A),\\
    &\lambda_s = \text{tanh}(\hat{S^i_{\nu}}) w_s^{\nu},\\
    &\lambda_d = \text{tanh}(\hat{S^i_{\nu}}) w_d^{\nu},\\
    &\tilde{\Theta}(S^i) \sim \mathcal{N}\Big(\rho^{\mu}(\dot{\Theta}^{\mu}(S^i)), \rho^{\sigma}(\dot{\Theta}^{\sigma}(S^i))\Big),\\
\end{split}
\end{align}
where $\tilde\Lambda^{\mu}_\text{filtered}$ and $\tilde\Lambda^{\sigma}_\text{filtered}$ are created following Equations (\ref{eq:uncertain_attentions}) and (\ref{eq:filtered_attention}).
Since the distribution of the output remains identical to the VGAT models, UA-FMCI-VGAT models can be created directly from VGAT models without the need to train the network again.

\section{Experiments}
We performed experiments on both spatial and spatio-temporal GCN tasks, which include the social trading analysis task in which the model predicts the trading behavior of socially connected investors, and the skeleton-based human action recognition task in which the model classifies human actions based on sequences of human body pose graphs. In the following, we describe these tasks and the conducted experiments.

\subsection{Social Trading Analysis}
The task of social trading analysis is commonly approached as a spatial problem in which the connections of the investors are static, and the model attempts to predict the trading behavior of an investor given the social connections between different investors that allow them to exploit private information for their own benefit \cite{rantala2019investors_exploit_information}, which is known as insider trading.
Insider trading is usually illegal, and the ability to capture such events with machine learning can help in ongoing investigations or even be a reason to open such an investigation.
However, since the distribution of private information is mostly done in a personal manner, it is difficult to capture such events \cite{baltakys2023ego}.
Deep learning fits well into the task of dealing with the incomplete inputs with complex dependencies and, for this reason, different researchers have used neural networks to analyze interconnected stocks \cite{qian2024gnn_stock, chang2009stocknn, pang2020stocknn} or investors \cite{baltakys2023ego} to predict their behavior or find suspicious transactions.

Baltakys et al. \cite{baltakys2023ego} are the first to predict investor trading activity based on their social connections in an insider network with the Finnish board membership dataset, presented in the same paper.
They use GCN and GAT architectures for one-shot analysis of the egocentric sub-graphs, with connections aggregated over the whole time-period of the dataset. They observe that actual social connections yield higher predictability compared to randomly reshuffled links, where investors’ empirical neighbors are replaced by other actual investors in the network. This suggests that social links between insiders are utilized for information transfer. In their paper, GAT models provide better detection accuracy than GCN models, which both outperform classical methods such as Logistic Regression and Support Vector Machine on this dataset. 
Deep Learning has also been used for some other social and financial tasks.
DeepInf \cite{qiu2018deepinf} uses GCN and GAT models to create graph embeddings of the social connections network to predict social influence in the network. 
Eagle \cite{shi2023eagle} uses a GNN model to detect tax evasion activity in a heterogeneous graph.
GCNs and Graph Autoencoders were used in \cite{munozcancino2023gnncredit} to predict credit trustworthiness based on a social interactions graph.

We used the public version of the Finnish board membership dataset \cite{baltakys2023ego} for our experiments involving spatial GCNs.
We compare the performance of the baseline GCN and GAT models to that of the proposed VGCN and VGAT models using the F1 metric. 
Following a similar process as the one proposed in \cite{oleksiienko2024uaab3dmot}, we initialize VGCN and VGAT models with the corresponding pretrained GCN and GAT models. These models are denoted as IVGCN and IVGAT.
The dataset is split into subsets created based on the prediction task, frequency and trading direction.
The Lead-lag task is used for predicting the future investor action based on the current actions of neighbors, and the Simultaneous task is used for predicting the current investor action based on the current neighbor actions.
The frequency can be either daily (D) or weekly (W) and the direction is either Buy or Sell.

\begin{figure}[!ht]
\centering
    \includegraphics[width=\linewidth]{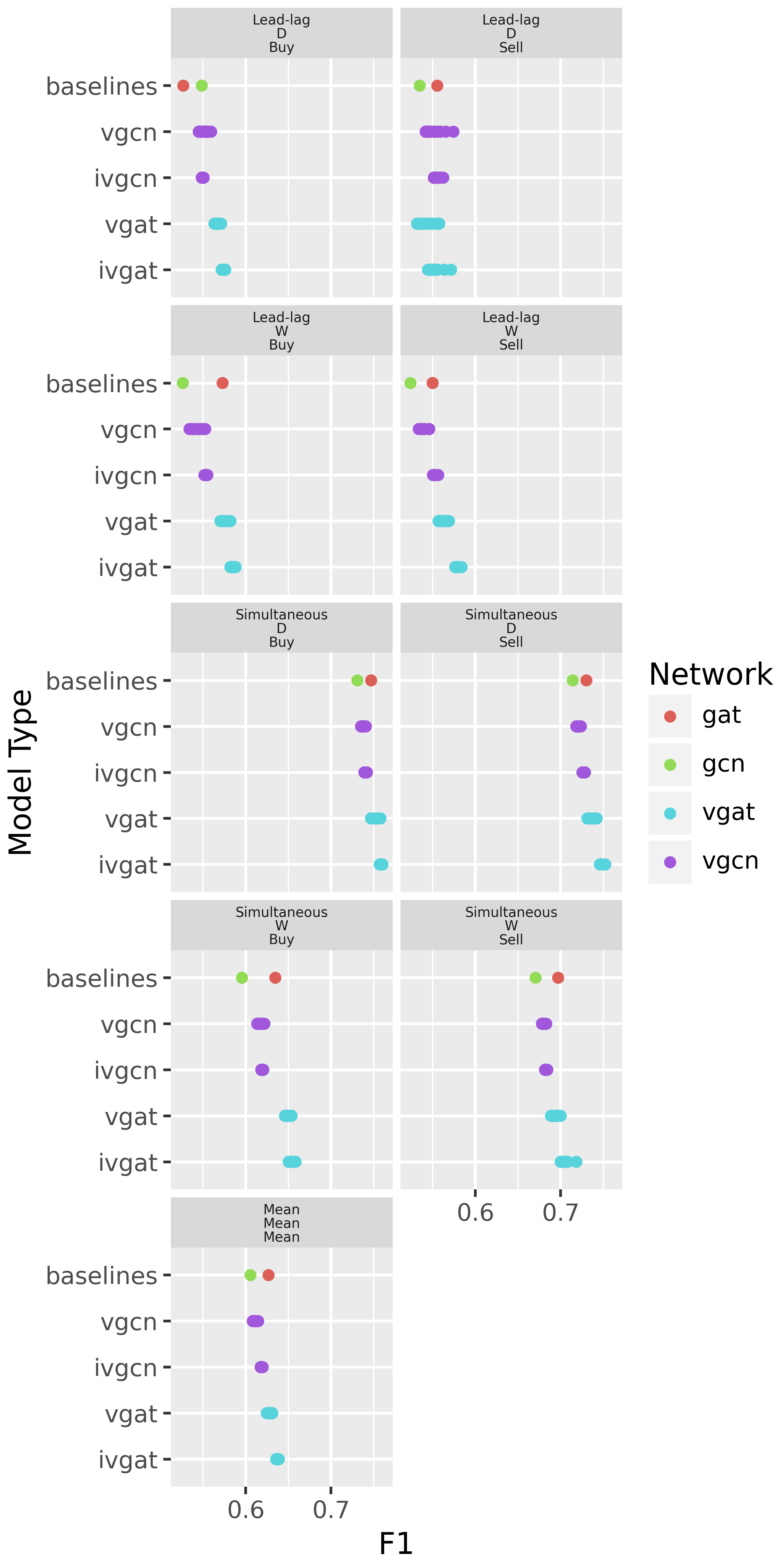}
    \caption{Results of GCN, GAT and variational versions of GCN and GAT models trained on the Finnish board membership dataset \cite{baltakys2023ego}. The Variational GCN (VGCN) and Variational GAT (VGAT) models are trained from scratch, while the IVGCN and IVGAT models are initialized with a pretrained GCN or a pretrained GAT model, respectively. The results are grouped by the subset, and mean results are given in a separate subfigure. For each model type, 20 best models with different hyperparameters are displayed.}
    \label{fig:results-insider}
\end{figure}

We experimented with different model hyperparameters for VGCNs, including the position of the activation functions, number of training samples for each of the inputs and the use of global variance.
Figure \ref{fig:results-insider} shows the evaluation results on all subsets of the Finnish board membership dataset, as well as the average results.
Each of the models is trained with 10 different random seeds and the results are averaged across seeds. Variational models also average their results over multiple tests of the same model due to their stochastic nature. The 20 best models are shown for the variational networks.
The GAT models outperform the GCN models in almost all subsets.
The same dynamic can be seen for VGAT versus VGCN, which also outperform the corresponding baselines.
Finally, the IVGAT and IVGCN outperform the VGAT and VGCN models, respectively.

\begin{table}[!ht]
        \caption{Comparison between base variational and uncertainty-aware models on Finnish board membership dataset.}
        \label{tab:ua_results}
        \centering
        \begin{tabular}{l|c}
            \toprule
            \textbf{Method} & Mean F1  \\ \hline
            IVGAT & 0.635  \\
            UA-EA-VGAT & 0.637 \\
            UA-FMCI-VGAT & 0.632 \\
            \bottomrule
        \end{tabular}
\end{table}

\begin{figure}[!ht]
\centering
    \includegraphics[width=\linewidth]{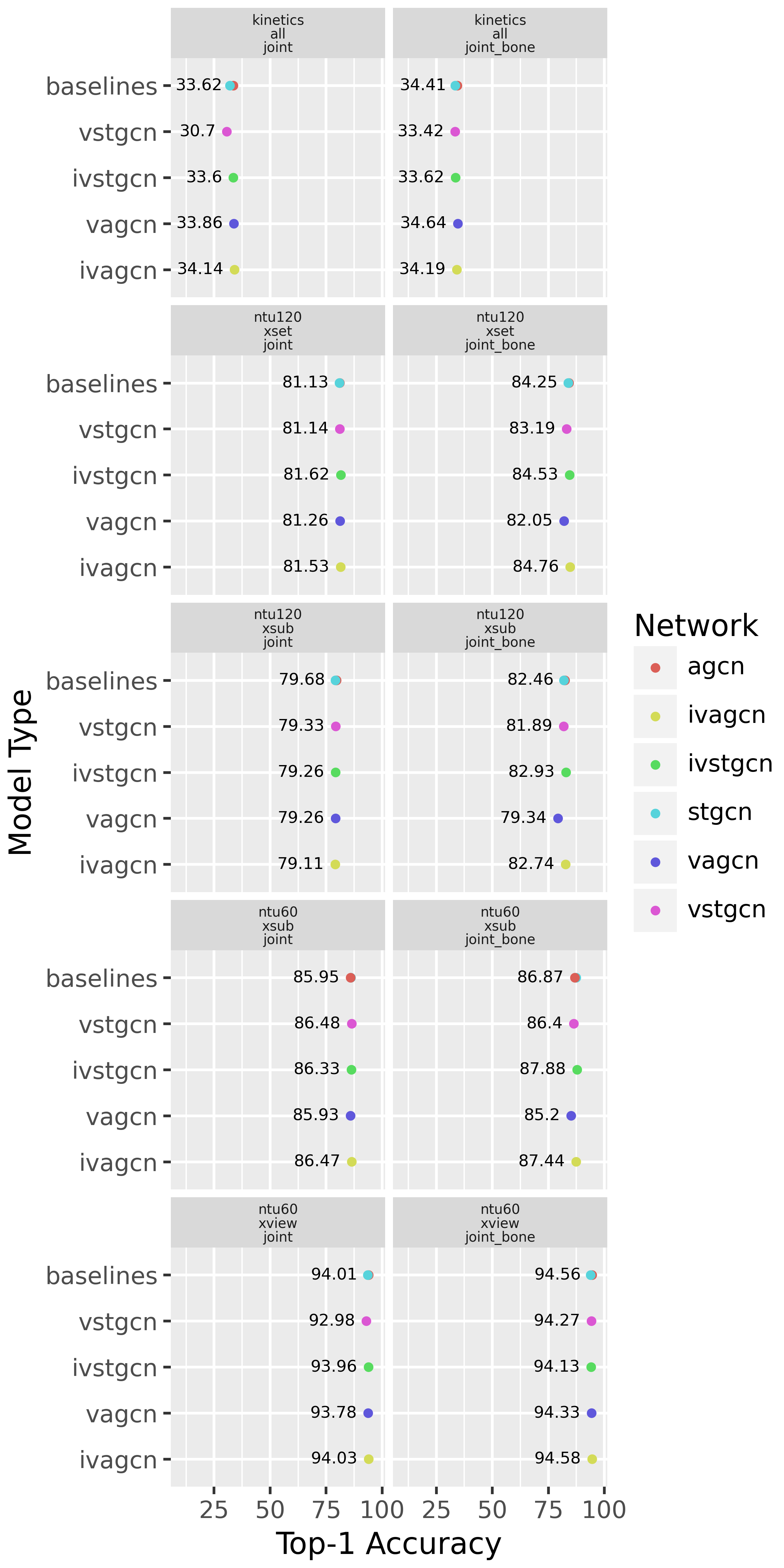}
    \caption{Results of ST-GCN, AGCN and variational versions of ST-GCN and ST-GAT models trained on the NTU-60 \cite{shahroudy2016ntu60}, NTU-120 \cite{liu2020ntu120}, and Kinetics \cite{will2017kinetics400} datasets. The Variational ST-GCN (VSTGCN) and Variational AGCN (VAGCN) models are trained from scratch, while the IVSTGCN and IVAGCN models are initialized with a pretrained ST-GCN or a pretrained AGCN model, respectively. The results are grouped by the dataset and corresponding subsets.}
    \label{fig:results-har}
\end{figure}

We also tested Uncertainty-Aware models on the social trading analysis task and compared the performance of IVGAT, the Uncertainty-Aware Early Attention VGAT (UA-EA-VGAT), and the Uncertainty-Aware Fully Monte Carlo Integrated VGAT (UA-FMCI-VGAT).
Table \ref{tab:ua_results} shows the mean F1 score over all subsets of the Finnish board membership datasets for the tested models.
The UA-EA-VGAT model shows a slight improvement over the IVGAT model, while UA-FMCI-VGAT leads to a slight performance drop.

\subsection{Skeleton-based Human Action Recognition}
Skeleton-based human action recognition is a spatio-temporal task in which a video is processed with a pose estimation method to extract human body skeletons representing human body poses at each video frame.
This leads to a series of graphs, where the graph nodes are connected according to the human body anatomy, and the features of the nodes change in time based on the movement of the human body.
To classify the action in a video, the method needs to process both the spatial data and their temporal variations, finding the relations between different joints in time.

Skeleton-based human action recognition is commonly approached with Spatio-temporal Graph Convolutional Networks.
ST-GCN \cite{yan2018stgcn} applies a spatial GCN layer and then aggregates spatial features with a temporal convolution in each of the ST-GCN blocks.
AGCN \cite{shi2019agcn} computes attention as a combination of a learnable matrix and a computed matrix from the input features.
TAGCN \cite{heidari2021tagcn} uses temporal attention to select the most informative skeletons and process only the needed parts of the video, reducing the computational complexity.
PST-GCN \cite{heidari2021pstgcn} progressively creates an architecture of an ST-GCN model.
ProtoGCN \cite{liu2024protogcn} applies prototype training to better discriminate between actions with similar joint trajectories.

The spatio-temporal models are evaluated on the NTU-60 \cite{shahroudy2016ntu60}, the NTU-120 \cite{liu2020ntu120}, and the Kinetics \cite{will2017kinetics400} datasets.
Performance is evaluated based on top-1 accuracy on the different subsets of each of the datasets.
The NTU-60 dataset has a cross-view (xview) subset and a cross-subject (xsub) subset, while the NTU-120 dataset has a cross-setup (xset) subset and a cross-subject (xsub) subset. 
The models can process either only the joint data of input skeletons, or both the joint and the bone data. 
Figure \ref{fig:results-har} shows the obtained experimental results obtained by applying the baselines ST-GCN \cite{yan2018stgcn} and AGCN \cite{shi2019agcn}, and their variational versions VST-GCN, IVST-GCN, VAGCN and IVAGCN.
The plots are grouped by the dataset, subset and skeleton type.
The variational networks provide a slight improvement in model accuracy in addition to providing the ability to estimate the model and attention uncertainties.

\begin{figure}[!ht]
\centering
    \includegraphics[width=\linewidth]{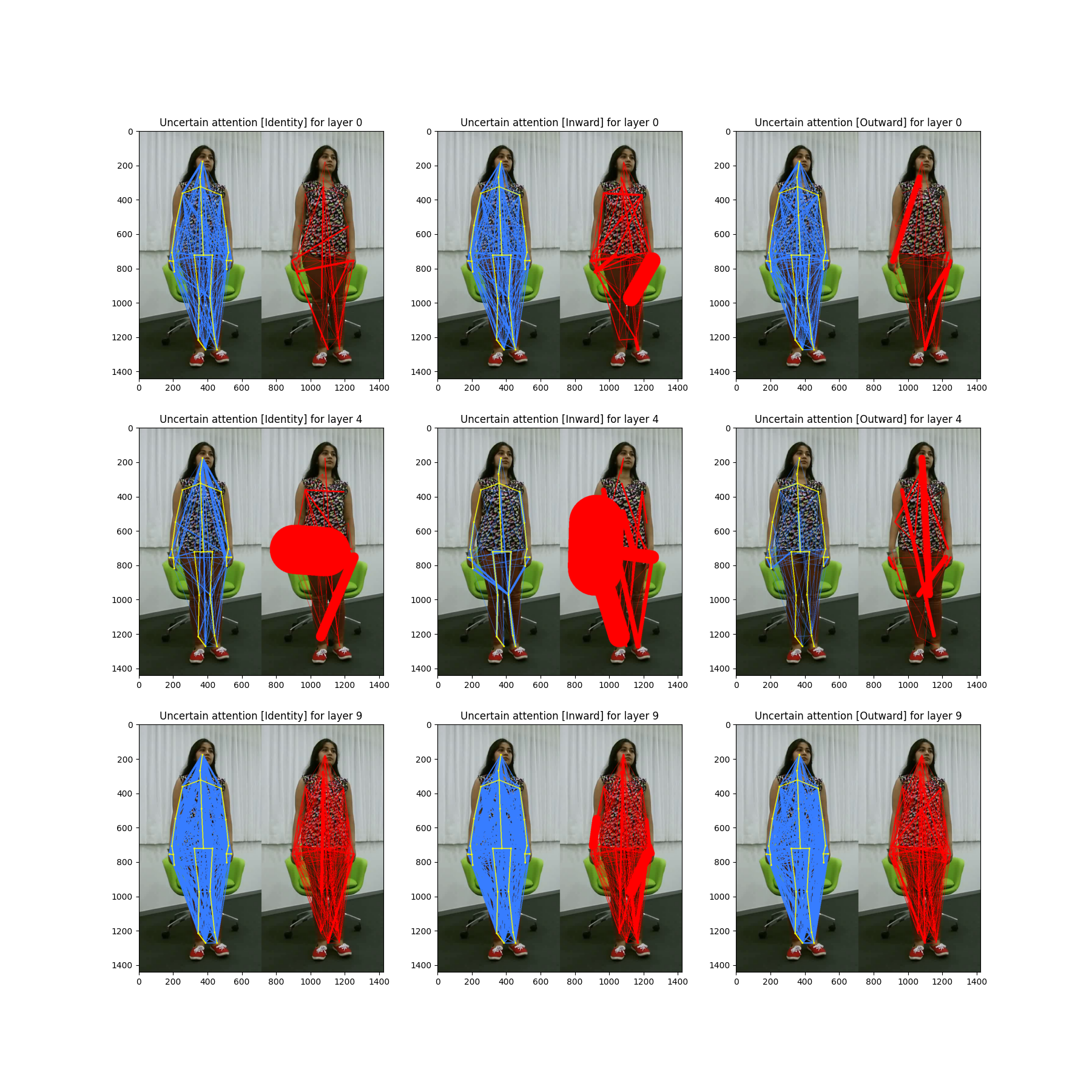}
    \caption{An example of layer-wise attention graphs for a single NTU-60 input sequence, displayed on top of the first frame in the video. The width of the blue lines represents the value of attention between the pair of joints, while the yellow lines represent the skeleton graph. The width of the red lines represents the uncertainty in the attention value, and is scaled to the same range, as the range of attention values. Each layer has a mean (left) and a variance (right) for each of the partitions of the attention matrices.}
    \label{fig:attention-har}
\end{figure}

An example of the final attentions in a VAGCN model is shown in Figure \ref{fig:attention-har}. For each layer and for each partition, we show the computed expectation and uncertainty in the attentions of the model.
Among the trained networks, spatio-temporal models show a higher level of uncertainty in attention than the spatial models, which can also be influenced by the difficulty of the problem.

Both VST-GCN and VAGCN do not consider all the possible activation places as in VGCN and VGAT because these models are much bigger and only the best activation options are chosen for the experiments on skeleton-based Human Action Recognition, based on the results from the smaller VGCN and VGAT networks.
This design choice is not dictated by the nature of ST-GCN models and can easily be augmented for other tasks.
Our implementation\footnote{\url{https://gitlab.au.dk/maleci/skeleton/skeleton-based-action-recognition}} of VSTGCN and VAGCN models supports all activation options, as well as the implementation\footnote{\url{https://github.com/iliiliiliili/insider-influence}} of VGCN and VGAT models.


\section{Conclusion}
In this paper, we proposed variational versions of spatial and spatio-temporal Graph Convolutional Networks, which allow estimating uncertainty in model outputs and attentions, as well as improving the model accuracy.
We evaluated the spatial GCN, GAT, VGCN and VGAT models on the Finnish board membership dataset for the social trading analysis task, and the spatio-temporal ST-GCN, AGCN, VSTGCN, VAGCN on the NTU-60, NTU-120 and Kinetics datasets for skeleton-based human action recognition task.
Variational models show a noticeable performance improvement for the financial task and a slight improvement for the human action recognition task.
The estimated uncertainties in the model outputs can be used to select if an additional verification of the results is needed. Both the output and the attention uncertainties can be used to improve model explainability and to identify the statistically most important links.

\bibliographystyle{IEEEbib}
\bibliography{bibliography.bib}

\end{document}